\documentclass[letterpaper, 10 pt, conference]{ieeeconf}  

\IEEEoverridecommandlockouts                              

\makeatletter

\def\ps@IEEEtitlepagestyle{%
	\def\@oddfoot{\mycopyrightnotice}%
	\def\@evenfoot{}%
}
\def\mycopyrightnotice{%
	{\footnotesize  978-1-4799-6773-5/14/\$31.00 \textcopyright2017 Crown\hfill}
	\gdef\mycopyrightnotice{}
}



\makeatletter
\newcommand*\titleheader[1]{\gdef\@titleheader{#1}}
\AtBeginDocument{%
	\let\st@red@title\@title%
	\def\@title{%
		\bgroup\normalfont\large\centering\@titleheader\par\egroup
		\vskip1.5em\st@red@title}
}
\makeatother

\usepackage{amsmath} 
\usepackage{amssymb}  
\usepackage{mathtools}
\usepackage{siunitx}

\usepackage{multirow}
\usepackage[table,xcdraw]{xcolor}
\usepackage{graphicx}

\usepackage{todonotes}
\usepackage[nolist]{acronym}
\usepackage{flushend}
\begin{acronym}
\acro{GPU}{Graphics Processing Units}
\acro{DNN}{Deep Neural Networks}
\acro{CNN}{Convolutional Neural Networks}
\acro{RNN}{Recurrent Neural Networks}
\acro{GNN}{Graph Neural Networks}
\acro{GCN}{Graph Convolutional Networks}
\acro{GRU}{Gated Rectifier Unit}
\acro{LSTM}{Long Short-Term Memory}
\acro{KF}{Kalman Filters}
\acro{BEV}{Bird's Eye View}
\acro{GAN}{Generative Adversarial Networks}
\acro{MTP}{Multi-Modal Trajectory Prediction}
\acro{MLP}{Multilayer Perceptron}
\acro{MAE}{Mean Absolute Error}
\acro{MSE}{Mean Squared Error}
\acro{FDE}{Final Displacement Error}
\acro{DKM}{Deep Kinematic Models}
\acro{FFW-ASP}{Feed-Forward Action-Space Prediction}
\acro{SSP-ASP}{Self-Supervised Action-Space Prediction}
\end{acronym}

\DeclareMathOperator{\sgn}{sgn}

\title{\LARGE \bf
Self-Supervised Action-Space Prediction for Automated Driving
}

\titleheader{\vspace{-45pt} \scriptsize ©2021 IEEE. Personal use of this material is permitted. Permission from IEEE must be obtained for all other uses, in any current
or future media, including reprinting/republishing this material for advertising or promotional purposes, creating new collective
works, for resale or redistribution to servers or lists, or reuse of any copyrighted component of this work in other works.}

\author{Faris Janjo\v{s}$^{1}$, Maxim Dolgov$^{1}$, and J. Marius Z\"ollner$^{2}$
	\thanks{$^{1}$ Robert Bosch GmbH, Corporate Research, Advanced Autonomous Systems, 71272 Renningen, Germany. {\tt\small \{faris.janjos, maxim.dolgov\}@de.bosch.com}}%
	\thanks{$^{2}$ Research Center for Information Technology (FZI), 76131 Karlsruhe, Germany.
		{\tt\small zoellner@fzi.de}}%
}

\begin{document}

\maketitle
\thispagestyle{empty}
\pagestyle{empty}

\begin{abstract}
Making informed driving decisions requires reliable prediction of other vehicles' trajectories. In this paper, we present a novel learned multi-modal trajectory prediction architecture for automated driving. It achieves kinematically feasible predictions by casting the learning problem into the space of accelerations and steering angles -- by performing action-space prediction, we can leverage valuable model knowledge. Additionally, the dimensionality of the action manifold is lower than that of the state manifold, whose intrinsically correlated states are more difficult to capture in a learned manner. For the purpose of action-space prediction, we present the simple \ac{FFW-ASP} architecture. Then, we build on this notion and introduce the novel \ac{SSP-ASP} architecture that outputs future environment context features in addition to trajectories. A key element in the self-supervised architecture is that, based on an observed action history and past context features, future context features are predicted prior to future trajectories. The proposed methods are evaluated on real-world datasets containing urban intersections and roundabouts, and show accurate predictions, outperforming state-of-the-art for kinematically feasible predictions in several prediction metrics.
\vspace{10pt} 
\end{abstract}

\vspace{-3pt}
\section{INTRODUCTION}

Trajectory prediction is a crucial component of automated driving stacks. Its task is to digest traffic context information given in the form of raw sensor data or intermediate representations, in order to infer goals and intended motions of other traffic participants. Accurately predicting trajectories of surrounding agents is a prerequisite for downstream planning components, whose job is to navigate the autonomous vehicle reasonably and safely by executing high-level behavior plans and performing lower-level trajectory planning and vehicle control~\cite{paden2016survey}. 

Predicting trajectories of human-driven vehicles shares complexities of understanding and predicting human motion in general~\cite{rudenko2020human}. Historically, most of the interest in human driving behavior and vehicle trajectory prediction has been shown in the context of driver assistance systems that employed classical robotics methods such as \ac{KF}~\cite{lefevre2014survey}, which perform well for short-term predictions but fail to capture intent-motivated long-term behavior. More information collected aboard vehicles, availability of large datasets, and the computing power of \ac{GPU} have driven the use of \ac{DNN} to achieve longer prediction horizons, as well as addressing full self-driving~\cite{tampuu2020survey}.

\begin{figure}
	\centering
	\includegraphics[width=0.48\textwidth]{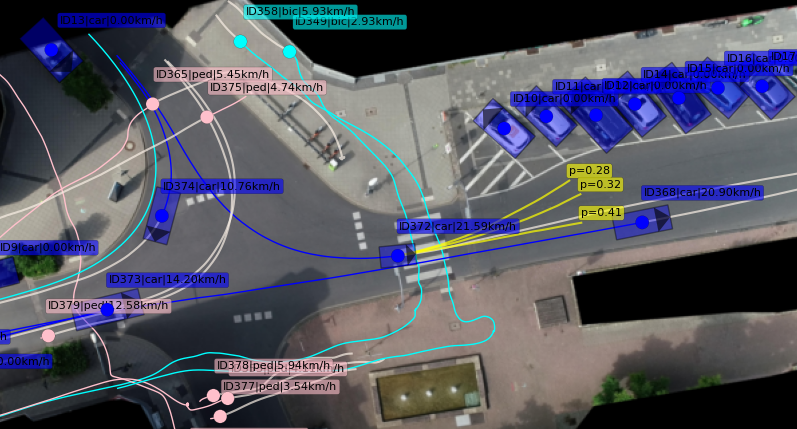}
	\caption{Multi-modal trajectory prediction (yellow) trained on the inD dataset \cite{bock2019ind}, using the \ac{SSP-ASP} approach}
	\label{fig:example_prediction}
\end{figure}

Despite the large body of research, addressing vehicle trajectory prediction still remains a difficult problem to be solved especially for reasons such as complexity of understanding subtle social cues in multi-agent interactions and necessity of predicting multi-modal trajectories (see Fig.~\ref{fig:example_prediction}). Furthermore, many works neglect kinematic vehicle constraints in the framing of the learning problem~\cite{cui2019deep}. Instead, the networks are tasked with capturing kinematic motion models in addition to the social and topological context by feeding inputs containing full vehicle states (position, heading, velocity, acceleration, etc.) and predicting future positions or waypoints as outputs.

In this work, we avoid these issues by constraining the learning problem to the action space and design architectures that benefit from such formulation. Our contributions are:
\begin{itemize}
\item We present a novel action-space learning paradigm, where the learning of motions is constrained purely to the action space of controlled accelerations and steering angles, both from the input and output sides. Position values or state values are then inferred via explicit kinematic motion models.
\item A self-supervised action-based prediction architecture with several novel concepts:
\begin{itemize}
\item Prediction of context features that are used for trajectory prediction
\item Reconstruction of action inputs using context features, helping the network learn a manifold sufficient for prediction of both features and trajectories
\item Long-term prediction (longer than 3\si{\second}) by chaining two predicted segments, which can be regarded as a conceptual middle ground between autoregressive/single-step approaches and one-shot/multi-step approaches.
\end{itemize} 
\end{itemize}
\section{RELATED WORK}\label{sec:related_work}

Learned models such as \ac{DNN}s trained on large datasets of potentially high-dimensional data (real or simulated) have shown capability at tackling vehicle trajectory prediction. They perform combined reasoning about static environment information (e.g. geometric road structure) and dynamic multi-agent interaction to extract sufficient latent statistics necessary to infer reasonable future trajectories. Among these models, different deep architectures as well as their combinations are: 
\begin{itemize}
	\item \textit{\ac{CNN}}~\cite{lecun2015deep} capture planar or spatial information from multi-channel images to construct locational visual features relevant to a scene.
	\item \textit{\ac{RNN}}~\cite{lecun2015deep} naturally model the sequential nature of the prediction problem, since consecutive points of a trajectory are propagated in a recurrent manner. 
	\item \textit{\ac{GNN}}~\cite{scarselli2008graph} encode geometric structure and interactions among different agents into an attention-based graph. By having lower-dimensional inputs than \ac{CNN}s, large reductions in the number of model weights can be made. A special class are \ac{GCN}, which employ so-called graph convolutions to extract richer topological representations than vanilla \ac{GNN}s~\cite{liang2020learning}.
\end{itemize}
In relation to this distinction, a categorization by different input modalities to \textit{raster}-, \textit{polyline}-, or \textit{sensor-based} approaches can be made. \textit{Raster-based} approaches~\cite{cui2019multimodal}, \cite{cui2019deep}, \cite{strohbeckmultiple}, \cite{chai2019multipath}, \cite{bansal2018chauffeurnet}, \cite{phan2020covernet} take in \ac{BEV} images of an agent's environment with elements such as road lanes and bounding boxes of other perceived agents. By doing so, the \ac{CNN} network can extract local spatial information from a single representational domain. Even though notions of distance are represented adequately, a large computational effort in training is needed to extract relevant features. In contrast, \textit{polyline-based} approaches~\cite{gao2020vectornet}, \cite{zhao2020tnt}, \cite{liang2020learning}, \cite{khandelwal2020if} use \ac{GNN}s that accept polylines or vectors representing surrounding roads and nearby agents. Another approach is predicting directly from sensor data such as LiDAR point clouds~\cite{casas2018intentnet}, \cite{casas2020implicit}, \cite{rhinehart2018r2p2}, \cite{rhinehart2019precog}, \cite{liang2020pnpnet}. Such \textit{sensor-based} approaches can perform object detection in addition to prediction, and propagate perception uncertainty throughout the pipeline. However, they require a very large computational effort in training and usually are not robust to sensor set changes.

More fundamentally, prediction approaches can be categorized by different learning paradigms that answer the question whether the prediction yields identical outputs given the same deterministic input data. Many \textit{supervised-learning} approaches have consistent outputs given deterministic data -- they either directly regress a trajectory \cite{cui2019multimodal}, \cite{cui2019deep}, or model uncertainty by predicting trajectories as sequences of means and covariances of positions, thus modeling the trajectories as uncorrelated distributions~\cite{chai2019multipath}, \cite{djuric2020uncertainty}, \cite{djuric2020multinet}. Similarly, some approaches improve prediction by performing classification over a predefined trajectory set~\cite{chai2019multipath}, \cite{phan2020covernet} or a target position set~\cite{zhao2020tnt}. In contrast, \textit{latent variable} approaches output random values w.r.t. the same deterministic data, by sampling a latent distribution while inferring trajectory candidates~\cite{rhinehart2018r2p2}, \cite{rhinehart2019precog}, \cite{rhinehart2018deep}, \cite{tang2019multiple}, \cite{casas2020implicit}. For example~\cite{rhinehart2019precog} defines prediction in this setting as trajectory forecasting, a class of imitation learning with non-interaction. Such approaches usually incorporate probabilistic latent variable priors that capture uncertainty as a set of non-interpretable random variables, or leverage \ac{GAN}~\cite{huang2019diversity}. However, their major drawback is the reliance on sequential sampling, which accumulates errors in each step. A notable exception is~\cite{casas2020implicit}, offering a high-dimensional continuous latent model employing \ac{GNN} encoder and decoder networks. 

Furthermore, approaches can be distinguished into two classes based on how a trajectory is generated: (i) \textit{one-shot} prediction approaches~\cite{chai2019multipath}, \cite{cui2019multimodal}, \cite{casas2020implicit} that output an entire trajectory and (ii) \textit{single-step}, autoregressive approaches that build a trajectory sequentially \cite{rhinehart2018r2p2}, \cite{rhinehart2019precog}, \cite{tang2019multiple}. According to how they model multi-agent interaction, most existing works perform prediction for each agent individually. Examples of these \textit{single-agent} approaches are~\cite{cui2019multimodal}, \cite{cui2019deep}, \cite{chai2019multipath}, \cite{strohbeckmultiple}. Exceptions exist that consider the set of agents \textit{jointly}, such as~\cite{tang2019multiple}, \cite{casas2020implicit}, \cite{rhinehart2019precog}. For example,~\cite{tang2019multiple} offers a discrete latent variable model with recurrent encoder and decoder networks, enabling joint prediction of all actors' trajectories in a scene.

Because the agents' true intentions are not known, trajectory predictions are inherently multi-modal. When it comes to modeling multi-modality, many \textit{deterministic-mode} approaches directly predict multiple candidate trajectories along with their probabilities~\cite{cui2019multimodal}, \cite{cui2019deep}. Similarly, \textit{Gaussian-mode} approaches represent a trajectory as a weighted mixture (each element is a candidate trajectory) of multi-variate Gaussians time-wise and predict means and variances~\cite{chai2019multipath}, \cite{hong2019rules}. Alternatively, previously mentioned \textit{sampling-based} approaches draw many samples from their decoder networks to obtain a notion of how likely a certain mode is. A notable class are \textit{heuristics-based} approaches, which either train an additional network to predict an anchor trajectory from a set of heuristically chosen trajectories~\cite{chai2019multipath}, \cite{phan2020covernet}, or use the road structure to infer viable target positions or regions~\cite{khandelwal2020if}, \cite{zhao2020tnt}. 

Many of the presented approaches can be abstracted to a duo of generic encoder/decoder networks (potentially with an additional prior network) with deterministic or random outputs, where backbone encoders compress the scene into features from which decoders infer trajectories. This assumes a complete reliance on learning to model physically feasible trajectory generation for the agent(s) of interest, which can be accurately described by a kinematic motion model instead. Such approaches opt to learn this model by reasoning about correlations among individual state variables, which in turn produces strong requirements on diversity in training data. A notable exception is~\cite{cui2019deep}, where a network learns action variables that are propagated through a bicycle model to infer trajectories. However, it achieves this by mapping states to actions, which assumes learning an inverse motion model instead. Similar approach that learns actions for prediction is~\cite{schulz2019learning}, which employs a large set of manually designed situational features as inputs.

In our approach, we argue that combining learned and physical models, in addition to yielding kinematically feasible trajectories, improves robustness by uncoupling modelable aspects from the learning task. To this end, we move the learning into the action space, by transforming action inputs to action outputs, and then use kinematic models to reconstruct trajectories. More fundamentally, action inputs in data are direct results of driver commands, as apposed to vehicle positions that are results of applied actions, and thus have the potential to model driving behavior more closely. 

Learning action outputs is prevalent in end-to-end approaches such as~\cite{tampuu2020survey}, \cite{bojarski2016end}, \cite{xiao2020action}. An advantage of using action outputs for the prediction problem is that they allow to reason more closely about causal effects of actions on internal network representations. Practically, this can be achieved by principles of self-supervision, which is used in the context of representation learning for applications such as automatic label generation~\cite{kolesnikov2019revisiting}, \cite{goyal2019scaling}. However, it finds applications in imitation learning as well~\cite{agrawal2016learning}, \cite{xiao2020action}. Here, self-supervision usually entails learning forward and inverse transformations concurrently~\cite{agrawal2016learning}. For example, if a certain action applied to a forward model results in an output described by generic features, it should be possible to reconstruct this applied action by feeding two consecutive features to an inverse model, whose job is to output actions. Inspired by these principles, we design a self-supervised trajectory prediction architecture, presented in Sec.~\ref{subsec:methods_self-sup_asp}.
\section{PROBLEM DESCRIPTION}\label{sec:problem_desc}

In general, trajectory prediction of human-driven vehicles can be framed in an imitation learning setting of modeling a distribution of trajectories $\mathbf{Y}$ given data $\mathcal{D}$
\vspace{-2pt}
\begin{equation}\label{eq:il_prediction}
	\mathbf{Y} \sim P(\mathbf{Y}|\mathcal{D})\ .
\vspace{-2pt}
\end{equation}
Data $\mathcal{D}$ can include prior trajectory information of the vehicle for which prediction is done (prediction-ego), sensor data acquired from the environment, or any kind of stored knowledge such as map information. It is an established approach to introduce a categorical hidden variable -- mode, and predict trajectories deterministically or as a probability distribution, conditioned on the mode. Then, in view of the \textit{deterministic-mode} approaches mentioned in Sec.~\ref{sec:related_work}, the problem in \eqref{eq:il_prediction} can be simplified by predicting $m$ modes of the distribution, i.e. deterministic trajectory $Y_m$ and probability $p_m$ pairs .

Trajectories $Y$ can be defined in several ways -- a vehicle's position trajectory is a $T$-step sequence of planar coordinates $X_{1:T}$, where each $X_t$ is represented by $[x, y]_t$. Similarly, we choose to define a full-state trajectory by the orientation and velocity in addition to positions; it is a sequence $S_{1:T}$, where $S_t$ is $[x, y, \theta, v]_t$, and $X_{1:T} \subset S_{1:T}$. Additionally, we define an action trajectory $A_{1:T}$ as a sequence of controlled accelerations $a_t$ and steering angles $\delta_t$, where $A_t$ is $[a, \delta]_t$. These action values fully capture the behavior of the driver along the time horizon $1:T$ in a given driving situation.

More precisely, we are interested in inferring the future position trajectory $X_{1:T}$ for the prediction ego vehicle, given data $\mathcal{D}$ that includes either past $X_{-T:0}$, $S_{-T:0}$, or $A_{-T:0}$, as well as generic environment observations $O_{-T:0}$. Each $O_t$ can contain observations in the form of LiDAR or \ac{BEV} rasterized images of the vehicle's surroundings, including road structure and other agents (vehicles, cyclists, pedestrians). In our implementation, we opt for rasterized \ac{BEV} representations, as displayed in Fig.~\ref{fig:rasters}.

\subsection{Action-space prediction}\label{subsec:asp}

In action prediction, the task is to predict future driving actions $A_{1:T}$. Therefore, learned action-space prediction can be defined as a generic learned mapping of past actions (without past positions and states) and generic observations to future actions
\vspace{-2pt}
\begin{equation}\label{eq:asp}
\{A_{-T:0}, O_{-T:0}\} \mapsto A_{1:T}\ .
\end{equation}
For simplicity, \eqref{eq:asp} only shows uni-modal action prediction, however a multi-modal extension to $m$ action modes $A_{1:T,m}, p_m$ is straightforward. 

Future position values $X_{1:T}$ are included in $S_{1:T}$, which is obtained by iteratively propagating actions $A_{1:T}$ through a kinematic motion model $f$
\vspace{-2pt}
\begin{equation}
S_{t+1} = f(S_t, A_t)\ .
\vspace{-2pt}
\end{equation}
Kinematic models of reasonable complexity sufficiently approximate true vehicle motion. They are virtually always valid and the cases where they are wrong (e.g. skidding) are rare and easy to detect. We opt for the bicycle model \cite{kong2015kinematic}, and perform the full state update in the following manner
\vspace{-2pt}
\begin{align}\label{eq:bicycle}
\begin{split}
x_{t+1} &= x_t + v_t\cos(\theta_t+\beta_t)\Delta T\ ,\\
y_{t+1} &= y_t + v_t\sin(\theta_t+\beta_t)\Delta T\ ,\\
\theta_{t+1} &= \theta_t + \frac{v_t}{l_r} \sin(\beta_t)\Delta T\ ,\\
v_{t+1} &= v_t + a_t\Delta T\ ,\\
\beta_t &= \tan^{-1}\left(\frac{l_r}{l_r+l_r} \tan\delta_t\right)\ .
\end{split}
\vspace{-2pt}
\end{align}
Here, $\Delta T$ is the sampling time, $\beta$ is the angle between the center-of-mass velocity and the longitudinal axis of the car, while $l_f$ and $l_r$ are distances from the center of mass to the front and rear wheels. Similarly, we define an inverse bicycle model as a state-to-action mapping. Due to overdetermined \eqref{eq:bicycle}, we choose to obtain the actions via $v$ and $\theta$
\vspace{-3pt}
\begin{align}\label{eq:inv_bicycle}
\begin{split}
a_t &= \frac{v_{t+1} - v_t}{\Delta T}\ ,\\
\delta_t &= \sgn\left(\frac{\theta_{t+1} - \theta_t}{\bar{v}_t}\right)\arctan\left(\frac{l_f + l_r}{\sqrt{(\frac{\bar{v}_t}{\theta_{t+1} - \theta_t})^2 - l^2_r}}\right),
\end{split} 
\vspace{-3pt}
\end{align}  
where $\bar{v}_t = \frac{v_t+v_{t+1}}{2}$. Identical inverse model is used in \cite{schulz2019learning}.

\subsection{Learned mappings}\label{subsec:mappings}

In terms of modeling kinematic characteristics and while ignoring generic observations, the previously presented \textit{action-to-action} mapping in \eqref{eq:asp} brings several benefits to the learning problem, contrary to learning \textit{state-to-position}~\cite{cui2019multimodal}, \textit{state-to-state}~\cite{strohbeckmultiple}, or \textit{state-to-action}~\cite{cui2019deep} mappings:
\begin{itemize}
	\item In a \textit{state-to-position} or \textit{state-to-state} mapping, the network has to implicitly capture a kinematic motion model (e.g. \eqref{eq:bicycle}) while reasoning about the correlation among individual states. To capture a valid model, a large number of diverse motions need to be present in the data (and not be underrepresented), such as driving straight, slight/sharp turns, U-turns, hard braking, etc.
	\item A \textit{state-to-action} mapping, used in \ac{DKM} approach \cite{cui2019deep}, incorporates an explicit forward motion model (e.g. \eqref{eq:bicycle}) to obtain states from actions. However, the network still has to implicitly capture an inverse model (e.g. \eqref{eq:inv_bicycle}) from sufficiently diverse data.
	\item An \textit{action-to-action} mapping fully captures kinematic characteristics and reduces requirements on the motion representation in the data. Importantly, it is of lower dimensionality than \textit{state-to-state} for instance, and makes capturing correlations between action variables easier.
\end{itemize}


Action-to-action prediction requires acceleration and steering angle values in tracked data, otherwise they can be inferred from states via inverse models such as \eqref{eq:inv_bicycle}. Examples of datasets where accelerations are provided are \cite{bock2019ind}, \cite{krajewski2020round}.

\section{ACTION-SPACE PREDICTION METHODS}\label{sec:methods}
\vspace{-2pt}
In this section, we present architectures that employ action-space prediction. We introduce the following notation:
\begin{itemize}
	\vspace{-1pt}
\item{\makebox[1cm]{$\tau_0$} is the past $T$-step time interval $[-T:0]$}
\item{\makebox[1cm]{$\tau_1$} is the future $T$-step time interval $[1:T+1]$}
\item{\makebox[1cm]{$\tau_2$} is the future $T$-step time interval $[T+2:2T+2]$}
\item{\makebox[1cm]{$\hat{}$} denotes predicted values}
\item{\makebox[1cm]{$*$} denotes ground truth values}
\end{itemize}

\subsection{Feed-forward action-space prediction: \ac{FFW-ASP}}\label{subsec:methods_ff_asp}
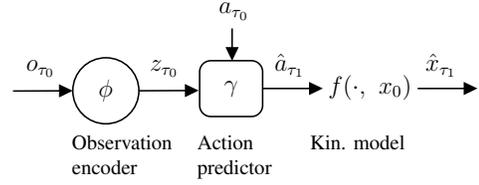
\begin{figure}[t]
	\centering
	\vspace{2pt}
	\scalebox{0.78}{\tikzset{every picture/.style={line width=0.75pt}} 

\begin{tikzpicture}[x=0.75pt,y=0.75pt,yscale=-1,xscale=1]

\draw   (97,126.33) .. controls (97,114.53) and (106.57,104.97) .. (118.37,104.97) .. controls (130.17,104.97) and (139.73,114.53) .. (139.73,126.33) .. controls (139.73,138.13) and (130.17,147.7) .. (118.37,147.7) .. controls (106.57,147.7) and (97,138.13) .. (97,126.33) -- cycle ;
\draw    (58.33,126.5) -- (94,126.35) ;
\draw [shift={(97,126.33)}, rotate = 539.75] [fill={rgb, 255:red, 0; green, 0; blue, 0 }  ][line width=0.08]  [draw opacity=0] (8.93,-4.29) -- (0,0) -- (8.93,4.29) -- cycle    ;
\draw    (139.73,126.33) -- (162.72,126.23) -- (175.4,126.18) ;
\draw [shift={(178.4,126.17)}, rotate = 539.75] [fill={rgb, 255:red, 0; green, 0; blue, 0 }  ][line width=0.08]  [draw opacity=0] (8.93,-4.29) -- (0,0) -- (8.93,4.29) -- cycle    ;
\draw   (179,114.09) .. controls (179,110.1) and (182.24,106.87) .. (186.23,106.87) -- (212.51,106.87) .. controls (216.5,106.87) and (219.73,110.1) .. (219.73,114.09) -- (219.73,135.77) .. controls (219.73,139.76) and (216.5,143) .. (212.51,143) -- (186.23,143) .. controls (182.24,143) and (179,139.76) .. (179,135.77) -- cycle ;
\draw    (200,86) -- (200,104) ;
\draw [shift={(200,107)}, rotate = 270] [fill={rgb, 255:red, 0; green, 0; blue, 0 }  ][line width=0.08]  [draw opacity=0] (8.93,-4.29) -- (0,0) -- (8.93,4.29) -- cycle    ;
\draw    (220,124.86) -- (255.67,124.84) ;
\draw [shift={(258.67,124.83)}, rotate = 539.96] [fill={rgb, 255:red, 0; green, 0; blue, 0 }  ][line width=0.08]  [draw opacity=0] (8.93,-4.29) -- (0,0) -- (8.93,4.29) -- cycle    ;
\draw    (320,124.86) -- (355.67,124.84) ;
\draw [shift={(358.67,124.83)}, rotate = 539.96] [fill={rgb, 255:red, 0; green, 0; blue, 0 }  ][line width=0.08]  [draw opacity=0] (8.93,-4.29) -- (0,0) -- (8.93,4.29) -- cycle    ;

\draw (97,129) node [anchor=north west][inner sep=0.75pt]   [align=left] { };
\draw (76.23,111.88) node  [font=\large]  {${\textstyle o_{\tau _{0}}}$};
\draw (156.23,111.88) node  [font=\large]  {${\displaystyle z_{\tau _{0}}}$};
\draw (201.23,74.88) node  [font=\large]  {$a_{\tau _{0}}$};
\draw (237.23,109.88) node  [font=\large]  {$\hat{a}_{\tau _{1}}$};
\draw (199.37,124.93) node  [font=\large]  {$\gamma $};
\draw (118.37,126.33) node  [font=\large]  {$\phi $};
\draw (95,153.67) node [anchor=north west][inner sep=0.75pt]   [align=left] {Observation\\encoder};
\draw (176,153.67) node [anchor=north west][inner sep=0.75pt]   [align=left] {Action\\predictor};
\draw (289.23,123.88) node  [font=\large]  {${\textstyle f( \cdot ,\ x_{0})}$};
\draw (335.23,109.88) node  [font=\large]  {$\hat{x}_{\tau _{1}}$};
\draw (249,153.67) node [anchor=north west][inner sep=0.75pt]   [align=left] {Kin. model};
\draw (122,148.4) node [anchor=north west][inner sep=0.75pt]    {$$};

\end{tikzpicture}}
	\caption{\ac{FFW-ASP} architecture: encoder with parameters $\phi$ accepts a history of observations $o_{\tau_0}$ and generates features $z_{\tau_0}$. Then, a decoder with parameters $\gamma$ combines features $z_{\tau_0}$ with action history $a_{\tau_0}$ to generate future actions $\hat{a}_{\tau_1}$. Kinematic model $f(\cdot, x_0)$ reconstructs a position trajectory}
	\label{fig:ff_asp}
\end{figure}

The \ac{FFW-ASP} architecture, shown in Fig. \ref{fig:ff_asp}, realizes the mapping of past actions and observations to future actions (given in \eqref{eq:asp}) via the standard encoder/decoder architecture. It can be described by the following sequence of mappings
\vspace{-3pt}
\begin{align}
o_{\tau_0} &\xrightarrow{\phi} z_{\tau_0}\ ,\\
z_{\tau_0}, a_{\tau_0} &\xrightarrow{\gamma} \hat{a}_{\tau_1}\ ,\\
\hat{x}_{\tau_1} &= f(\hat{a}_{\tau_1}, x_0)\ ,
\vspace{-3pt}
\end{align}
where $\phi$ parameterizes an encoder network that maps past observations to features, and $\gamma$ parameterizes a decoder that maps features and past actions to future actions. Future positions are obtained from \eqref{eq:bicycle} via the kinematic model $f(\cdot)$ that requires the present position $x_0$ in addition to future actions $\hat{a}_{\tau_1}$, in order to reconstruct future positions $\hat{x}_{\tau_1}$.

The loss function penalizes the difference between predicted positions and ground truth positions. In our implementation, we opt for the Huber loss with a cut-off at $h=1.0$
\vspace{-12pt}
\begin{align}\label{eq:huber}
\begin{split}
&\Delta x_{\tau_1} = \hat{x}_{\tau_1} - x^*_{\tau_1}\ ,\\
\mathcal{L} = \Vert &\Delta x_{\tau_1} \Vert = \begin{cases}
\frac{1}{2} \Delta x^2_{\tau_1}, \quad &|\Delta x_{\tau_1}| < h\\
h (|\Delta x_{\tau_1}| - \frac{1}{2}h), \quad &\text{otherwise}\ .
\end{cases}
\end{split}
\vspace{-3pt}
\end{align}
This loss function incorporates predicted positions obtained via the fully-differentiable kinematic model, making backpropagation possible.

It is straightforward to extend this architecture to multi-modal predictions. Inspired by~\cite{cui2019multimodal}, the action decoder can be adapted to predict $M$ action trajectories $\hat{a}_{\tau_1, m}$ along with their probabilities $p_m$, where $m\in\{1,...,M\}$. In this case, the loss function is extended with a binary indicator function $I$ and a cross-entropy loss term
\begin{equation}\label{eq:multi_modal_reg_loss}
\mathcal{L}_{FFW} = \sum_{m=1}^{M}I_{m=m^\dagger}(\Vert \Delta x_{\tau_1, m} \Vert) - \frac{ \exp (p_{m^\dagger})}{\sum_m \exp(p_m)}\ ,
\end{equation}
where the indicator selects the mode closest to the ground-truth (denoted by $\dagger$), while the cross-entropy term rewards confidence in the chosen mode. The mode-to-ground-truth distance uses a combination of angle difference between the last points of the two trajectories and $L_2$ distance between each point \cite{cui2019multimodal}.

This simple feed-forward architecture capitalizes on the benefits of \textit{action-to-action} mappings from Sec.~\ref{subsec:asp}. In the next section, we describe the methodology of self-supervision in an action-based context, before presenting a self-supervised trajectory prediction architecture.

\subsection{Self-supervision background}\label{subsec:methods_ss_asp} 

The forward and inverse transformations in the context of self-supervision, previously mentioned in Sec.~\ref{sec:related_work}, have been used in robotic manipulation tasks~\cite{agrawal2016learning} to jointly learn mappings from visual features generated from images of objects to physical robot actions on the objects and vice-versa. This architecture has been adapted for end-to-end driving in~\cite{xiao2020action}, where a pre-trained self-supervised forward/inverse model outperforms pure end-to-end models that map images to driving actions. Furthermore, a categorization of different learned transformations is offered in~\cite{xiao2020action}, which we summarize here with one-step predictions.

\begin{enumerate}
\item \textbf{Behavioral cloning} models map an observation to latent features with a $\phi$-encoder, and map features to action output with a $\gamma$-decoder 
\vspace{-4pt}
\begin{equation}
o_t \xrightarrow{\phi} z_t \xrightarrow{\gamma} \hat{a}_{t}\ .
\vspace{-3pt}
\end{equation}
The loss is a difference between the label and the output
\vspace{-4pt}
\begin{equation}
\mathcal{L}_{BC} = \Vert a^*_{t} - \hat{a}_{t} \Vert\ .
\vspace{-5pt}
\end{equation}

\item \textbf{Inverse} models encode observations into features at two successive time-steps. Then, they learn a $\xi$-parameterized mapping from two consecutive features to an action, to reconstruct the applied action that resulted in a new observation 
\vspace{-2pt}
\begin{equation}
\begin{rcases}
o_t &\xrightarrow{\phi} z_t  \\
o_{t+1} &\xrightarrow{\phi} z_{t+1} 
\end{rcases}
\xrightarrow{\xi} \hat{a}_t\ ,
\vspace{-2pt}
\end{equation}
with a reconstruction loss 
\vspace{-4pt}
\begin{equation}\label{eq:inverse_model}
\mathcal{L}_{INV} = \Vert a^*_{t} - \hat{a}_{t}\Vert\ .
\vspace{-2pt}
\end{equation}
\item \textbf{Forward} models, parameterized by $\psi$, predict the new latent features given past features and action
\vspace{-3pt}
\begin{align}
\begin{rcases}\label{eq:predicted_future_feature}
o_t \xrightarrow{\phi} &z_t\\
&a_t
\end{rcases} &\xrightarrow{\psi} \hat{z}_{t+1}\ , \\
o_{t+1} &\xrightarrow{\phi} z_{t+1}\label{eq:enc_future_feature}\ .
\vspace{-3pt}
\end{align}
Optimizing only the feature mismatch between the predicted future features $\hat{z}_{t+1}$ and the encoded features $z_{t+1}$ can lead to networks outputting the trivial solution of zero features. Therefore, a regularization term is added in the form of the inverse model loss \eqref{eq:inverse_model} 
\vspace{-2pt}
\begin{equation}\label{eq:reg_fw_model}
\mathcal{L}_{FW} = \Vert z_{t+1} - \hat{z}_{t+1}\Vert + \Vert a^*_{t} - \hat{a}_{t}\Vert\ .
\vspace{-2pt}
\end{equation}
\end{enumerate}

The regularized forward model is preferable to pure behavioral cloning, since the networks gain a richer understanding of the interplay between actions and features. The model can reconstruct its inputs, as well as predict its internal representation of the environment (in the form of features), as opposed to predicting observations directly, which can be infeasible in the case of camera images for example.

\subsection{Self-supervised action-space prediction: \ac{SSP-ASP}}\label{subsec:methods_self-sup_asp} 
The previously presented forward model (with an inverse model regularizer) predicts only future latent features. In order to use it for trajectory prediction, we add an action predictor component parameterized by $\gamma$
\vspace{-2pt}
\begin{equation}
a_t, z_t, \{z_{t+1}, \hat{z}_{t+1}\} \xrightarrow{\gamma} \hat{a}_{t+1}\ .
\vspace{-2pt}
\end{equation}
This mapping predicts the future action $\hat{a}_{t+1}$ by combining the known past action $a_t$, encoded past latent feature $z_t$, and one of the future features $\{z_{t+1}, \hat{z}_{t+1}\}$. The difference is that $z_{t+1}$ is used at training time and obtained via the encoding $\phi$~\eqref{eq:enc_future_feature}, while $\hat{z}_{t+1}$ is used at inference time and obtained via the prediction $\psi$~\eqref{eq:predicted_future_feature}. With the additional action predictor, the total self-supervised action prediction loss extends the regularized forward model loss \eqref{eq:reg_fw_model} according to
\vspace{-2pt}
\begin{equation}\label{eq:total_loss}
\mathcal{L}_{} = \Vert z_{t+1} - \hat{z}_{t+1}\Vert + \Vert a_{t} - \hat{a}_{t}\Vert + \Vert a^*_{t+1} -  \hat{a}_{t+1}\Vert\ .
\vspace{-2pt}
\end{equation}
In summary, it incorporates the forward model that predicts future features (feature predictor), the inverse model regularizer that 'predicts' the past actions (action reconstructor), and the action predictor, predicting future actions. 

\begin{figure}[t]
	\centering
	\vspace{2pt}
	\scalebox{0.8}{\tikzset{every picture/.style={line width=0.75pt}} 

\begin{tikzpicture}[x=0.75pt,y=0.75pt,yscale=-1,xscale=1]

\draw   (97,126.33) .. controls (97,114.53) and (106.57,104.97) .. (118.37,104.97) .. controls (130.17,104.97) and (139.73,114.53) .. (139.73,126.33) .. controls (139.73,138.13) and (130.17,147.7) .. (118.37,147.7) .. controls (106.57,147.7) and (97,138.13) .. (97,126.33) -- cycle ;
\draw    (58.33,126.5) -- (94,126.35) ;
\draw [shift={(97,126.33)}, rotate = 539.75] [fill={rgb, 255:red, 0; green, 0; blue, 0 }  ][line width=0.08]  [draw opacity=0] (8.93,-4.29) -- (0,0) -- (8.93,4.29) -- cycle    ;
\draw    (139.73,126.33) -- (162.72,126.23) -- (175.4,126.18) ;
\draw [shift={(178.4,126.17)}, rotate = 539.75] [fill={rgb, 255:red, 0; green, 0; blue, 0 }  ][line width=0.08]  [draw opacity=0] (8.93,-4.29) -- (0,0) -- (8.93,4.29) -- cycle    ;
\draw   (179,114.09) .. controls (179,110.1) and (182.24,106.87) .. (186.23,106.87) -- (212.51,106.87) .. controls (216.5,106.87) and (219.73,110.1) .. (219.73,114.09) -- (219.73,135.77) .. controls (219.73,139.76) and (216.5,143) .. (212.51,143) -- (186.23,143) .. controls (182.24,143) and (179,139.76) .. (179,135.77) -- cycle ;
\draw    (200,86) -- (200,104) ;
\draw [shift={(200,107)}, rotate = 270] [fill={rgb, 255:red, 0; green, 0; blue, 0 }  ][line width=0.08]  [draw opacity=0] (8.93,-4.29) -- (0,0) -- (8.93,4.29) -- cycle    ;
\draw    (220,124.86) -- (255.67,124.84) ;
\draw [shift={(258.67,124.83)}, rotate = 539.96] [fill={rgb, 255:red, 0; green, 0; blue, 0 }  ][line width=0.08]  [draw opacity=0] (8.93,-4.29) -- (0,0) -- (8.93,4.29) -- cycle    ;
\draw   (97,250.33) .. controls (97,238.53) and (106.57,228.97) .. (118.37,228.97) .. controls (130.17,228.97) and (139.73,238.53) .. (139.73,250.33) .. controls (139.73,262.13) and (130.17,271.7) .. (118.37,271.7) .. controls (106.57,271.7) and (97,262.13) .. (97,250.33) -- cycle ;
\draw    (58.33,250.5) -- (94,250.35) ;
\draw [shift={(97,250.33)}, rotate = 539.75] [fill={rgb, 255:red, 0; green, 0; blue, 0 }  ][line width=0.08]  [draw opacity=0] (8.93,-4.29) -- (0,0) -- (8.93,4.29) -- cycle    ;
\draw    (139.73,250.33) -- (162.72,250.23) -- (175.4,250.18) ;
\draw [shift={(178.4,250.17)}, rotate = 539.75] [fill={rgb, 255:red, 0; green, 0; blue, 0 }  ][line width=0.08]  [draw opacity=0] (8.93,-4.29) -- (0,0) -- (8.93,4.29) -- cycle    ;
\draw   (179,238.09) .. controls (179,234.1) and (182.24,230.87) .. (186.23,230.87) -- (212.51,230.87) .. controls (216.5,230.87) and (219.73,234.1) .. (219.73,238.09) -- (219.73,259.77) .. controls (219.73,263.76) and (216.5,267) .. (212.51,267) -- (186.23,267) .. controls (182.24,267) and (179,263.76) .. (179,259.77) -- cycle ;
\draw    (200,210) -- (200,228) ;
\draw [shift={(200,231)}, rotate = 270] [fill={rgb, 255:red, 0; green, 0; blue, 0 }  ][line width=0.08]  [draw opacity=0] (8.93,-4.29) -- (0,0) -- (8.93,4.29) -- cycle    ;
\draw    (220,248.86) -- (255.67,248.84) ;
\draw [shift={(258.67,248.83)}, rotate = 539.96] [fill={rgb, 255:red, 0; green, 0; blue, 0 }  ][line width=0.08]  [draw opacity=0] (8.93,-4.29) -- (0,0) -- (8.93,4.29) -- cycle    ;
\draw   (296,185.09) .. controls (296,181.1) and (299.24,177.87) .. (303.23,177.87) -- (373.67,177.87) .. controls (377.66,177.87) and (380.9,181.1) .. (380.9,185.09) -- (380.9,206.77) .. controls (380.9,210.76) and (377.66,214) .. (373.67,214) -- (303.23,214) .. controls (299.24,214) and (296,210.76) .. (296,206.77) -- cycle ;
\draw    (307,157) -- (307,175) ;
\draw [shift={(307,178)}, rotate = 270] [fill={rgb, 255:red, 0; green, 0; blue, 0 }  ][line width=0.08]  [draw opacity=0] (8.93,-4.29) -- (0,0) -- (8.93,4.29) -- cycle    ;
\draw    (339,157) -- (339,175) ;
\draw [shift={(339,178)}, rotate = 270] [fill={rgb, 255:red, 0; green, 0; blue, 0 }  ][line width=0.08]  [draw opacity=0] (8.93,-4.29) -- (0,0) -- (8.93,4.29) -- cycle    ;
\draw    (367.9,139.8) -- (368.91,175) ;
\draw [shift={(369,178)}, rotate = 268.35] [fill={rgb, 255:red, 0; green, 0; blue, 0 }  ][line width=0.08]  [draw opacity=0] (8.93,-4.29) -- (0,0) -- (8.93,4.29) -- cycle    ;
\draw    (339,214) -- (339,232) ;
\draw [shift={(339,235)}, rotate = 270] [fill={rgb, 255:red, 0; green, 0; blue, 0 }  ][line width=0.08]  [draw opacity=0] (8.93,-4.29) -- (0,0) -- (8.93,4.29) -- cycle    ;

\draw (97,129) node [anchor=north west][inner sep=0.75pt]   [align=left] { };
\draw (76.23,111.88) node  [font=\large]  {${\textstyle o_{\tau _{1}}}$};
\draw (156.23,111.88) node  [font=\large]  {${\displaystyle z_{\tau _{1}}}$};
\draw (201.23,74.88) node  [font=\large]  {$z_{\tau _{0}}$};
\draw (237.23,109.88) node  [font=\large]  {$\hat{a}_{\tau _{0}}$};
\draw (199.37,124.93) node  [font=\large]  {$\xi $};
\draw (118.37,126.33) node  [font=\large]  {$\phi $};
\draw (95,153.67) node [anchor=north west][inner sep=0.75pt]   [align=left] {Observation\\encoder};
\draw (176,153.67) node [anchor=north west][inner sep=0.75pt]   [align=left] {Action\\reconstructor};
\draw (308.23,145.88) node  [font=\large]  {$a_{\tau _{0}}$};
\draw (97,253) node [anchor=north west][inner sep=0.75pt]   [align=left] { };
\draw (76.23,235.88) node  [font=\large]  {${\textstyle o_{\tau _{0}}}$};
\draw (156.23,235.88) node  [font=\large]  {${\displaystyle z_{\tau _{0}}}$};
\draw (201.23,198.88) node  [font=\large]  {$a_{\tau _{0}}$};
\draw (237.23,233.88) node  [font=\large]  {$\hat{z}_{\tau _{1}}$};
\draw (199.37,248.93) node  [font=\large]  {$\psi $};
\draw (118.37,250.33) node  [font=\large]  {$\phi $};
\draw (95,277.67) node [anchor=north west][inner sep=0.75pt]   [align=left] {Observation\\encoder};
\draw (176,277.67) node [anchor=north west][inner sep=0.75pt]   [align=left] {Feature\\predictor};
\draw (340.23,246.88) node  [font=\large]  {$\hat{a}_{\tau _{1}}$};
\draw (339.95,195.93) node  [font=\large]  {$\gamma $};
\draw (290,263.67) node [anchor=north west][inner sep=0.75pt]   [align=left] {Action predictor};
\draw (340.23,145.88) node  [font=\large]  {$z_{\tau _{0}}$};
\draw (365.23,122.88) node  [font=\large]  {$\{z_{\tau _{1}} ,\hat{z}_{\tau _{1}}\}$};

\end{tikzpicture}}
	\caption{\ac{SSP-ASP} \textit{training} architecture: encoder with parameters $\phi$ accepts observations at two consecutive intervals $o_{\tau_0}$ and $o_{\tau_1}$ separately, generates features $z_{\tau_0}$ and $z_{\tau_1}$, and then the action reconstructor $\xi$ and feature predictor $\psi$ reconstruct past actions $\hat{a}_{\tau_0}$ and predict future features $\hat{z}_{\tau_1}$, respectively. The action predictor uses past actions $a_{\tau_0}$ and consecutive features $z_{\tau_0}$ and $z_{\tau_1}$ (or $\hat{z}_{\tau_1}$) to predict future actions $\hat{a}_{\tau_1}$. Kinematic transformation is omitted for clarity.}
	\label{fig:self-sup_train_asp}
\end{figure}
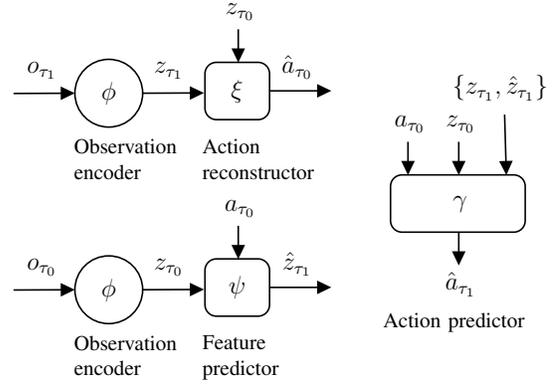

The complete procedure to obtain the training loss \eqref{eq:total_loss} is visualized in Fig.~\ref{fig:self-sup_train_asp} for $T$-step intervals. First, we encode observations from past and future intervals $o_{\tau_0}$ and $o_{\tau_1}$ separately into features $z_{\tau_0}$ and $z_{\tau_1}$. Then, we reconstruct the past action by generating $\hat{a}_{\tau_0}$ while simultaneously predicting future features $\hat{z}_{\tau_1}$. Finally, we predict future actions $\hat{a}_{\tau_1}$ using $a_{\tau_0}$, $z_{\tau_0}$, and $z_{\tau_1}$ (in training) or $\hat{z}_{\tau_1}$ (during inference). In practice, we run the action predictor with both encoded and predicted features in training -- we use both outputs in the loss, in order to avoid a distribution mismatch during inference. For inference, the architecture in Fig. \ref{fig:self-sup_train_asp} reduces to encoding only past observation $o_{\tau_0}$ (since $o_{\tau_1}$ is not available), predicting features $\hat{z}_{\tau_1}$, and then predicting future actions $\hat{a}_{\tau_1}$ using $a_{\tau_0}$, $z_{\tau_0}$, and $\hat{z}_{\tau_1}$. Finally, since we are interested in multi-modal position prediction over a $T$-step interval as opposed to a single time-step, \eqref{eq:total_loss} becomes
\vspace{-2pt}
\begin{align}\label{eq:total_mm_loss}
\begin{split}
\mathcal{L}_{SSP} &= w_1\Vert a_{\tau_0} - \hat{a}_{\tau_0}\Vert + w_2\Vert z_{\tau_1} - \hat{z}_{\tau_1}\Vert +\\ w_3&\sum_{m=1}^{M}I_{m=m^\dagger}(\Vert \Delta x_{\tau_1, m} \Vert) - w_4\frac{ \exp (p_{m^\dagger})}{\sum_m \exp(p_m)}\ .
\end{split}
\vspace{-2pt}
\end{align}
Here, the weights $\lbrace w_i\rbrace_{i=1}^4$ penalize action reconstruction, feature mismatch, multi-modal regression, and classification.

An important aspect of the self-supervised action prediction problem is its time-interval formulation. It enables us to introduce a distinction between \textit{single-segment} prediction, and a long-term, \textit{multi-segment} prediction.
\subsubsection{Single-segment prediction}
We have so far considered the same $T$-step time interval for all the past and future values. In this setup, we predict a single segment of values of interest, exemplified in Fig. \ref{fig:self-sup_train_asp}, with the constraint of same time interval length for the past and future. The duration can be regarded as a design choice. We opt for 3\si{\second} segment length since we assume that [-3:0]\si{\second} history is long enough to capture relevant scene context, and [0:3]\si{\second} future is a meaningful, albeit not long-term, prediction interval.
\subsubsection{Multi-segment prediction}
\begin{figure}[t]
	\centering
	\vspace{2pt}
	\scalebox{.8}{\tikzset{every picture/.style={line width=0.75pt}} 

\begin{tikzpicture}[x=0.75pt,y=0.75pt,yscale=-1,xscale=1]

\draw   (92,123.33) .. controls (92,111.53) and (101.57,101.97) .. (113.37,101.97) .. controls (125.17,101.97) and (134.73,111.53) .. (134.73,123.33) .. controls (134.73,135.13) and (125.17,144.7) .. (113.37,144.7) .. controls (101.57,144.7) and (92,135.13) .. (92,123.33) -- cycle ;
\draw    (134.73,122.33) -- (157.72,122.23) -- (170.4,122.18) ;
\draw [shift={(173.4,122.17)}, rotate = 539.75] [fill={rgb, 255:red, 0; green, 0; blue, 0 }  ][line width=0.08]  [draw opacity=0] (8.93,-4.29) -- (0,0) -- (8.93,4.29) -- cycle    ;
\draw   (174,111.09) .. controls (174,107.1) and (177.24,103.87) .. (181.23,103.87) -- (207.51,103.87) .. controls (211.5,103.87) and (214.73,107.1) .. (214.73,111.09) -- (214.73,132.77) .. controls (214.73,136.76) and (211.5,140) .. (207.51,140) -- (181.23,140) .. controls (177.24,140) and (174,136.76) .. (174,132.77) -- cycle ;
\draw   (215,168.09) .. controls (215,164.1) and (218.24,160.87) .. (222.23,160.87) -- (252.11,160.87) .. controls (256.1,160.87) and (259.33,164.1) .. (259.33,168.09) -- (259.33,189.77) .. controls (259.33,193.76) and (256.1,197) .. (252.11,197) -- (222.23,197) .. controls (218.24,197) and (215,193.76) .. (215,189.77) -- cycle ;
\draw    (176.33,179.94) -- (212,179.79) ;
\draw [shift={(215,179.77)}, rotate = 539.75] [fill={rgb, 255:red, 0; green, 0; blue, 0 }  ][line width=0.08]  [draw opacity=0] (8.93,-4.29) -- (0,0) -- (8.93,4.29) -- cycle    ;
\draw    (149.23,122.28) -- (149.33,219.67) -- (237.33,219.67) -- (237.33,198.67) ;
\draw [shift={(237.33,195.67)}, rotate = 450] [fill={rgb, 255:red, 0; green, 0; blue, 0 }  ][line width=0.08]  [draw opacity=0] (8.93,-4.29) -- (0,0) -- (8.93,4.29) -- cycle    ;
\draw    (193.67,178.86) -- (193.97,143) ;
\draw [shift={(194,140)}, rotate = 450.49] [fill={rgb, 255:red, 0; green, 0; blue, 0 }  ][line width=0.08]  [draw opacity=0] (8.93,-4.29) -- (0,0) -- (8.93,4.29) -- cycle    ;
\draw    (311.67,178.86) -- (311.97,143) ;
\draw [shift={(312,140)}, rotate = 450.49] [fill={rgb, 255:red, 0; green, 0; blue, 0 }  ][line width=0.08]  [draw opacity=0] (8.93,-4.29) -- (0,0) -- (8.93,4.29) -- cycle    ;
\draw    (237.33,121.85) -- (237.33,158.67) ;
\draw [shift={(237.33,161.67)}, rotate = 270] [fill={rgb, 255:red, 0; green, 0; blue, 0 }  ][line width=0.08]  [draw opacity=0] (8.93,-4.29) -- (0,0) -- (8.93,4.29) -- cycle    ;
\draw    (53.73,123.33) -- (76.72,123.23) -- (89.4,123.18) ;
\draw [shift={(92.4,123.17)}, rotate = 539.75] [fill={rgb, 255:red, 0; green, 0; blue, 0 }  ][line width=0.08]  [draw opacity=0] (8.93,-4.29) -- (0,0) -- (8.93,4.29) -- cycle    ;
\draw    (269.23,122.28) -- (269.33,219.67) -- (357.33,219.67) -- (357.33,198.67) ;
\draw [shift={(357.33,195.67)}, rotate = 450] [fill={rgb, 255:red, 0; green, 0; blue, 0 }  ][line width=0.08]  [draw opacity=0] (8.93,-4.29) -- (0,0) -- (8.93,4.29) -- cycle    ;
\draw    (259,179.86) -- (332.45,179.37) ;
\draw [shift={(335.45,179.35)}, rotate = 539.62] [fill={rgb, 255:red, 0; green, 0; blue, 0 }  ][line width=0.08]  [draw opacity=0] (8.93,-4.29) -- (0,0) -- (8.93,4.29) -- cycle    ;
\draw   (335,168.09) .. controls (335,164.1) and (338.24,160.87) .. (342.23,160.87) -- (372.11,160.87) .. controls (376.1,160.87) and (379.33,164.1) .. (379.33,168.09) -- (379.33,189.77) .. controls (379.33,193.76) and (376.1,197) .. (372.11,197) -- (342.23,197) .. controls (338.24,197) and (335,193.76) .. (335,189.77) -- cycle ;
\draw   (291,111.09) .. controls (291,107.1) and (294.24,103.87) .. (298.23,103.87) -- (324.51,103.87) .. controls (328.5,103.87) and (331.73,107.1) .. (331.73,111.09) -- (331.73,132.77) .. controls (331.73,136.76) and (328.5,140) .. (324.51,140) -- (298.23,140) .. controls (294.24,140) and (291,136.76) .. (291,132.77) -- cycle ;
\draw    (331.73,121.33) -- (354.72,121.23) -- (367.4,121.18) ;
\draw [shift={(370.4,121.17)}, rotate = 539.75] [fill={rgb, 255:red, 0; green, 0; blue, 0 }  ][line width=0.08]  [draw opacity=0] (8.93,-4.29) -- (0,0) -- (8.93,4.29) -- cycle    ;
\draw    (378.73,180.33) -- (401.72,180.23) -- (414.4,180.18) ;
\draw [shift={(417.4,180.17)}, rotate = 539.75] [fill={rgb, 255:red, 0; green, 0; blue, 0 }  ][line width=0.08]  [draw opacity=0] (8.93,-4.29) -- (0,0) -- (8.93,4.29) -- cycle    ;
\draw    (215,121.86) -- (288.45,121.37) ;
\draw [shift={(291.45,121.35)}, rotate = 539.62] [fill={rgb, 255:red, 0; green, 0; blue, 0 }  ][line width=0.08]  [draw opacity=0] (8.93,-4.29) -- (0,0) -- (8.93,4.29) -- cycle    ;
\draw    (357.33,120.85) -- (357.33,157.67) ;
\draw [shift={(357.33,160.67)}, rotate = 270] [fill={rgb, 255:red, 0; green, 0; blue, 0 }  ][line width=0.08]  [draw opacity=0] (8.93,-4.29) -- (0,0) -- (8.93,4.29) -- cycle    ;

\draw (72,109) node  [font=\large]  {${\textstyle o_{\tau _{0}}}$};
\draw (151.23,108.88) node  [font=\large]  {${\displaystyle z_{\tau _{0}}}$};
\draw (233.23,106.88) node  [font=\large]  {$\hat{z}_{\tau _{1}}$};
\draw (194.37,121.93) node  [font=\large]  {$\psi $};
\draw (113.37,123.33) node  [font=\large]  {$\phi $};
\draw (236.95,178.93) node  [font=\large]  {$\gamma $};
\draw (180.23,164.88) node  [font=\large]  {$a_{\tau _{0}}$};
\draw (298.23,161.88) node  [font=\large]  {$\hat{a}_{\tau _{1}}$};
\draw (398.23,161.88) node  [font=\large]  {$\hat{a}_{\tau _{2}}$};
\draw (350.23,106.88) node  [font=\large]  {$\hat{z}_{\tau _{2}}$};
\draw (356.95,178.93) node  [font=\large]  {$\gamma $};
\draw (311.37,121.93) node  [font=\large]  {$\psi $};

\end{tikzpicture}}
	\caption{\ac{SSP-ASP} two-segment \textit{inference} architecture: encoder with parameters $\phi$ accepts observations $o_{\tau_0}$ and then alternately, future features $\hat{z}_{\tau_1}$ and $\hat{z}_{\tau_2}$, and actions $\hat{a}_{\tau_1}$ and $\hat{a}_{\tau_2}$ are generated. First, $\hat{z}_{\tau_1}$ and $\hat{a}_{\tau_1}$ are generated with the help of encoded features $z_{\tau_0}$, and then, $\hat{z}_{\tau_2}$ and $\hat{a}_{\tau_2}$ with the help of predicted features $\hat{z}_{\tau_1}$. Note how the action reconstructor $\xi$ from Fig. \ref{fig:self-sup_train_asp} is not needed in inference.}
	\label{fig:self-sup_test_asp}
\end{figure}
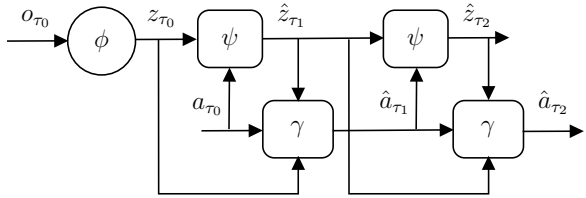
The self-supervised architecture makes it possible to chain several predicted segments and achieve long-term prediction. This can be done both in training and inference by additionally generating features and actions along the $\tau_2$ interval. The inference architecture is visualized in Fig. \ref{fig:self-sup_test_asp}. Here, the feature predictor $\psi$ is run once more to obtain $\hat{z}_{\tau_2}$, and action predictor $\gamma$ to obtain $\hat{a}_{\tau_2}$. For the $3s$ segment length it would constitute a prediction on the~[3:6]\si{\second} interval. The training case is similar to Fig. \ref{fig:self-sup_train_asp}, with $o_{\tau_1}$ and $o_{\tau_2}$ as inputs. The same procedure is carried out, obtaining encoded features ${z}_{\tau_1}$ and ${z}_{\tau_2}$, and predicting $\hat{z}_{\tau_2}$. Finally, $\hat{a}_{\tau_2}$ can be obtained either with the encoded $z_{\tau_2}$ or the predicted $\hat{z}_{\tau_2}$. In general, the multi-segment chaining can be performed with the same procedure for even more segments further into the future. In our implementation, we focus on two segments.

Multi-segment prediction requires non-trivial chaining of multi-modal trajectories. For example, the predicted action output along $\tau_1$ is multi-modal with $m$ modes, and it needs to be fed again into the action predictor $\gamma$ that accepts uni-modal inputs, and outputs multi-modal actions along $\tau_2$. Naturally, it is possible to feed $m$ modes individually and obtain $m^2$ chained trajectories at the output, requiring a large computational effort both in training and inference. A less costly alternative is feeding the highest probability trajectory within the multi-modal actions. Similarly, \textit{label matching} can be done -- the action mode closest to the label position trajectory can be chosen (after a kinematic transform of the actions), with the distance metric the same as in the regression component of the losses \eqref{eq:multi_modal_reg_loss} and \eqref{eq:total_mm_loss}. In this context, label matching can be regarded as a variant of guided teacher forcing \cite{williams1989learning} -- a method to accelerate training of \ac{RNN}s by replacing the outputs of earlier stages of the network with ground truth samples.

In the context of prediction, the self-supervised architectures in Fig. \ref{fig:self-sup_train_asp} and Fig. \ref{fig:self-sup_test_asp} improve on classic encoder-decoder setups exemplified by Fig. \ref{fig:ff_asp}. They achieve this by incorporating future observations in training, as opposed to only past observations, while at the same time predicting the future context and reconstructing the past actions. This joint learning helps all the networks form richer representations.

\section{RESULTS}

In this section, we describe the implementation details of the approach proposed in Sec.~\ref{sec:methods}. We specify the training procedure, the datasets that were used for training, and present the obtained results.

\subsection{Implementation}\label{subsec:results_impl}
\begin{figure}
	\centering
	\begin{minipage}[t]{0.38\columnwidth}
		\centering
		\includegraphics[width=1\textwidth]{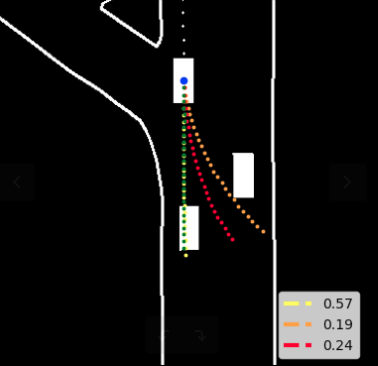}
	\end{minipage}
	\begin{minipage}[t]{0.38\columnwidth}
		\centering
		\includegraphics[width=.97\textwidth]{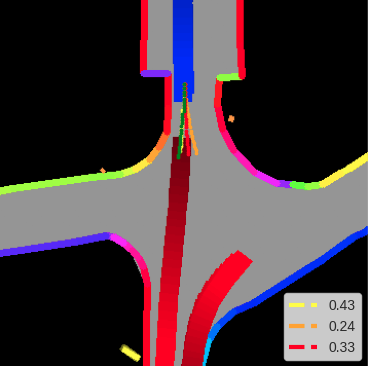}
	\end{minipage}
	\caption{Example predicted trajectories with ChauffeurNet (all channels superimposed) and \ac{MTP} rasters on inD~\cite{bock2019ind}; green trajectory indicates ground truth, while red, orange, and yellow trajectories are predicted modes}
	\label{fig:rasters}
\end{figure}

In Sec. \ref{sec:methods}, we have assumed generic observations as inputs to generic encoder networks that can reason with different input representations (as categorized in Sec.~\ref{sec:related_work}). In our implementation, we opt for \textit{raster-based} images and work with two variants visualized in Fig.~\ref{fig:rasters}. The first variant is a sparse black and white 12-channel 360x240 representation, similar to the ChauffeurNet representation~\cite{bansal2018chauffeurnet}. Here, individual channels contain the road boundary polylines, prediction-ego bounding box, prediction-ego trajectory history, and multiple snapshots (sampled at regular time-intervals) of bounding boxes of all traffic agents. The second variant is a 3-channel 360x360 RGB-image similar to the \ac{MTP} representation~\cite{cui2019multimodal}. Here, the road boundary polylines, drivable areas, and prediction-ego and traffic bounding boxes are overlaid on a single image, where different semantics have specific color values. We implemented this variant with the help of nuScenes devkit~\cite{nuscenes2019}.

The rasterized inputs are fed into ResNet18~\cite{he2016deep} encoder backbone networks returning a 512-sized feature vector, both for the case of feed-forward and self-supervised architectures. For the former, the $\gamma$ action predictor (Fig.~\ref{fig:ff_asp}) is realized as a 2-layer \ac{GRU} with 512 hidden units. For the self-supervised architecture (Fig.~\ref{fig:self-sup_train_asp}), we use the following networks: feature predictor $\psi$ and the action predictor $\gamma$ are also 2-layer \ac{GRU}s with 512 hidden units, whereas the action reconstructor $\xi$ is a \ac{MLP} with layer sizes \{256, 128, 64\}. The $\gamma$ and $\xi$ networks output acceleration and steering angle values in the range $[-1,1]$, which are then scaled to the range observed in the dataset.

In both feed-forward and self-supervised architectures we consider a 3\si{\second} history and observe two use-cases: predicting a 3\si{\second} and a 6\si{\second} trajectory. For the 6\si{\second} prediction, the trajectory is generated via two different prediction paradigms: one-shot prediction for the feed-forward architecture and chaining two 3\si{\second} predictions for the self-supervised architecture (with label matching to account for multi-modality).


\subsection{Datasets}
We evaluated our approach on two real-world datasets recorded on German roads: inD (urban intersections)~\cite{bock2019ind} and rounD (roundabouts)~\cite{krajewski2020round}. In total, they contain highly accurate drone-recorded trajectories of over 25000 agents of different classes, such as cars, trucks, buses, motorcycles, cyclists, and pedestrians. We performed prediction for all agent classes except cyclists and pedestrians. A drawback of the datasets is the lack of signalized intersections, which were not considered in our approach but can be easily integrated into the input representation. For example, the ChauffeurNet~\cite{bansal2018chauffeurnet} rasterization considers traffic light status as an additional channel.

Depending on the prediction horizon, we extracted all 3+3\si{\second} (3\si{\second} past and 3\si{\second} prediction) or 3+6\si{\second} (3\si{\second} past and 6\si{\second} prediction) trajectory snippets from the datasets, with a 0.6\si{\second} spacing. We split each dataset into training, validation, and testing subsets according to an 8:1:1 ratio, with each subset containing recordings taken at a different time of the day, thus ensuring no overlap. To generate the road boundaries polylines, we used the provided lanelet maps~\cite{poggenhans2018lanelet2}.

\subsection{Training setup}
The feed-forward and self-supervised architectures were trained on the accompanying losses \eqref{eq:multi_modal_reg_loss} and \eqref{eq:total_mm_loss}. For the self-supervised case, we used equal weighting for the loss components. Similarly to \eqref{eq:multi_modal_reg_loss}, each loss component is represented by the Huber loss function \eqref{eq:huber}. 

The self-supervised architecture offers great flexibility in different ways to pre-train the models. We experimented with training on only the action reconstruction and feature mismatch loss components from \eqref{eq:total_mm_loss} prior to the full loss, essentially implementing \eqref{eq:reg_fw_model}. We observed that this \textit{forward-model pre-training} improves performance since it enables the features to be learned to a certain degree before the actual prediction is performed. Additionally, in the 6\si{\second} prediction horizon case, we observed performance gains by loading and retraining a model used for 3\si{\second} prediction. 

All models were implemented in PyTorch~\cite{paszke2019pytorch} and trained on two NVIDIA GeForce GTX 1080 GPUs using the Adam optimizer~\cite{kingma2014adam}, a batch size of 32, and 10 epochs. The learning rate was set to $10^{-4}$ and reduced by a factor of five if the validation loss did not improve over three consecutive epochs.  In self-supervised prediction, we performed additional forward-model pre-training over 3 epochs.

\subsection{Performance}
We evaluated the performance of our \ac{FFW-ASP} (Sec.~\ref{subsec:methods_ff_asp}) and \ac{SSP-ASP} (Sec.~\ref{subsec:methods_ss_asp}) approaches on inD~\cite{bock2019ind} and rounD \cite{krajewski2020round} datasets. We benchmarked them against \ac{DKM}~\cite{cui2019deep}, another method that achieves kinematically feasible predictions, albeit by using a \textit{state-to-action} mapping introduced in Sec. \ref{subsec:mappings} compared to our proposed \textit{action-to-action} mapping. This method improves on the authors' prior work~\cite{cui2019multimodal}, whose multi-modality concept and rasterization we leverage. All methods were compared in both ChauffeurNet~\cite{bansal2018chauffeurnet} and \ac{MTP}~\cite{cui2019multimodal} rasterization setups. For a fair comparison, we used a ResNet18 encoder in all setups and performed 3\si{\second} and 6\si{\second} prediction based on a 3\si{\second} history (with 3 modes). As evaluation metrics, we calculated the~\ac{MAE}\footnote{\scriptsize{mean $L1$ distance across all time-steps}},~\ac{MSE}\footnote{\scriptsize{mean $L2$ distance across all time-steps}}, and~\ac{FDE}\footnote{\scriptsize{$L2$ distance at final time-step}} to the ground-truth on the testing datasets.

\begin{table}[ht]
	\centering
	\vspace{2pt}
	\scalebox{0.85}{\renewcommand\arraystretch{1.1}

\centering
\begin{tabular}{|c|c|ccc|ccc|} 
\hline
\multicolumn{2}{|c|}{\multirow{2}{*}{inD}}                                       & \multicolumn{3}{c|}{3\si{\second}}                                                              & \multicolumn{3}{c|}{6\si{\second}}                                                               \\
\multicolumn{2}{|c|}{}                                                           & \multicolumn{1}{c}{MAE}    & \multicolumn{1}{c}{MSE}    & \multicolumn{1}{c|}{FDE}    & \multicolumn{1}{c}{MAE}    & \multicolumn{1}{c}{MSE}    & \multicolumn{1}{c|}{FDE}     \\ 
\hline
\multirow{3}{*}{\begin{tabular}[c]{@{}c@{}}Chauffeur\\Net\end{tabular}} & DKM    & \multicolumn{1}{c}{$0.48$} & \multicolumn{1}{c}{$0.86$} & \multicolumn{1}{c|}{$2.01$} & \multicolumn{1}{c}{$1.27$} & \multicolumn{1}{c}{$7.44$} & \multicolumn{1}{c|}{$5.91$}  \\
& FFW-ASP & $0.27$                     & $0.33$                     & $1.23$                      & $\textbf{1.13} $                     & $\textbf{6.42}$                     & \multicolumn{1}{c|}{$5.45$}  \\
& SSP-ASP & $\textbf{0.21}$                     & $\textbf{0.24}$                     & $\textbf{0.98}$                      & $1.23 $                     & $6.52$                     & $\textbf{5.15}$                       \\ 
\hline
\multirow{3}{*}{MTP}                                                    & DKM    & $0.58$                     & $1.19$                     & $2.36$                      & $1.34$                     & $7.89$                     & $6.29$                       \\
& FFW-ASP & $0.30$                     & $0.39$                     & $1.41$                      & $1.19$                     & $6.78$                     & $5.72$                       \\
& SSP-ASP & $\textbf{0.22}$                     & $\textbf{0.25}$                     & $\textbf{1.04}$                      & $\textbf{1.15}$                     & $\textbf{6.08}$                     & $\textbf{5.02}$                       \\
\hline
		
\multicolumn{8}{c}{}\\

\hline
\multicolumn{2}{|c|}{\multirow{2}{*}{rounD}}                                       & \multicolumn{3}{c|}{3\si{\second}}                                                              & \multicolumn{3}{c|}{6\si{\second}}                                                               \\
\multicolumn{2}{|c|}{}                                                           & \multicolumn{1}{c}{MAE}    & \multicolumn{1}{c}{MSE}    & \multicolumn{1}{c|}{FDE}    & \multicolumn{1}{c}{MAE}    & \multicolumn{1}{c}{MSE}    & \multicolumn{1}{c|}{FDE}     \\ 
\hline
\multirow{3}{*}{\begin{tabular}[c]{@{}c@{}}Chauffeur\\Net\end{tabular}} & DKM    & \multicolumn{1}{c}{$0.29$} & \multicolumn{1}{c}{$0.29$} & \multicolumn{1}{c|}{$1.12$} & \multicolumn{1}{c}{$1.35$} & \multicolumn{1}{c}{$7.22$} & \multicolumn{1}{c|}{$5.55$}  \\
& FFW-ASP & $0.22$                     & $0.18$                     & $0.93$                      & $\textbf{1.12}$                     & $\textbf{6.17}$                     & \multicolumn{1}{c|}{$5.08$}  \\
& SSP-ASP & $\textbf{0.17}$                     & $\textbf{0.12}$                     & $\textbf{0.73}$                      & $1.34$                     & $6.44$                     & $\textbf{4.69}$                       \\ 
\hline
\multirow{3}{*}{MTP}                                                    & DKM    & $0.29$                     & $0.28$                     & $1.22$                      & $1.25$                     & $6.13$                     & $5.51$                       \\
& FFW-ASP & $0.22$                     & $0.18$                     & $0.94$                      & $\textbf{1.08} $                     & $\textbf{5.84}$                     & $4.76$                       \\
& SSP-ASP & $\textbf{0.17}$                     & $\textbf{0.12}$                     & $\textbf{0.74}$                      & $1.25$                     & $7.08$                     & $\textbf{4.61}$                       \\
\hline
\end{tabular}

}
	\caption{Results obtained on the inD \cite{bock2019ind} and rounD \cite{krajewski2020round} datasets}\vspace{-20pt}
	\label{tab:results_table}
\end{table}

The results are provided in Tab. \ref{tab:results_table}, showing that \ac{FFW-ASP} already brings improvements over \ac{DKM} on both datasets. Since the two approaches differ in the choice of the prediction head network (\ac{DKM} has a fully connected layer, while \ac{FFW-ASP} has a \ac{GRU}), we tested \ac{FFW-ASP} using a \ac{DKM} prediction head and still achieved improvements (omitted in the table). Furthermore, \ac{SSP-ASP} offers clear improvements to \ac{FFW-ASP} on both datasets for 3\si{\second} prediction, while for 6\si{\second} prediction the feed-forward architecture achieves similar results overall. This indicates that the method of chaining multi-modal trajectories, specific to the multi-segment variant, could potentially be improved. Nevertheless, \ac{SSP-ASP} achieves the lowest \ac{FDE} across all settings.



\section{CONCLUSION}

In this paper, we offered an action-based perspective into the problem of vehicle trajectory prediction. Results indicate that uncoupling modelable aspects from the learning problem, such as learning forward and inverse motion models, improves prediction performance while ensuring kinematic feasibility. As a main result, we proposed a novel self-supervised action-based architecture for prediction. By predicting future latent features and reconstructing past actions simultaneously, we capture the interplay between actions and their effects on the networks' internal representations. We extract more information by leveraging future observations in training, which we use to predict future latent features -- a feasible alternative to predicting observations. Finally, we achieve long-term prediction via a new paradigm: in contrast to one-shot or single-step prediction approaches, we obtain a trajectory by chaining individual trajectory segments.

In future work, we plan to replace the computationally intensive \ac{CNN} encoder with a lightweight \ac{GNN}, as this is currently the bottleneck of the self-supervised approach. We expect more robust feature representation and thus a performance improvement, since the network would not have to extract objects from pixels. Furthermore, we would like to analyze the effects of time segment length on prediction quality. It remains to be seen whether shorter or longer segments can capture more latent information in a scene and whether chaining more than two segments could potentially achieve prediction horizons longer than 6\si{\second}.




\bibliographystyle{IEEEtran}
\bibliography{references/bibliography}
\clearpage

\end{document}